\documentclass[11pt]{article}
\usepackage[a4paper]{geometry}
\usepackage{clic2017}
\usepackage{times}
\usepackage{url}
\usepackage{latexsym}
\usepackage{graphicx}

\usepackage{amsmath}

\usepackage{soul}
\usepackage[dvipsnames]{xcolor}

\title{NLP-CIC @ DIACR-Ita: POS and Neighbor Based Distributional Models for Lexical Semantic Change in Diachronic Italian Corpora\thanks{\ \ ``Copyright \textcopyright\
2020 for this paper by its authors. Use permitted under Creative Commons License Attribution 4.0 International (CC BY 4.0).''}}

  \author{ 
  Jason Angel \\
  CIC, Instituto Polit\'{e}cnico Nacional \\
  Mexico City, Mexico \\
  {\tt ajason08@gmail.com}
  \And
  Carlos A. Rodriguez-Diaz \\
  CIC, Instituto Polit\'{e}cnico Nacional \\
  Mexico City, Mexico \\
  {\tt amnet04@gmail.com}
  \AND
  Alexander Gelbukh \\
  CIC, Instituto Polit\'{e}cnico Nacional \\
  Mexico City, Mexico \\
  {\tt www.gelbukh.com}
  \And  
  Sergio Jimenez  \\
  Instituto Caro y Cuervo \\
  Bogota, Colombia \\
  {\tt  sergio.jimenez@caroycuervo.gov.co}   
  } 

\date{}

\begin{document}
\maketitle
\begin{abstract}
 
 We present our systems and findings on unsupervised lexical semantic change for the Italian language in the DIACR-Ita shared-task at EVALITA 2020. 
  The task is to determine whether a target word has evolved its meaning with time, only relying on raw-text from two time-specific datasets.
  We propose two models representing the target words across the periods to predict the changing words using threshold and voting schemes. Our first model solely relies on part-of-speech usage and an ensemble of distance measures. The second model uses word embedding representation to extract the neighbor's relative distances across spaces and propose ``the average of absolute differences" to estimate lexical semantic change. Our models achieved competent results, ranking third in the DIACR-Ita competition. Furthermore, we experiment with the $k\_neighbor$ parameter of our second model to compare the impact of using ``the average of absolute differences" versus the cosine distance used in \cite{hamilton-etal-2016-cultural}.
  



\end{abstract}


\section{Introduction}
Lexical semantic change has recently gained interest in the intersection of natural language processing and historical linguistics\footnote{see \url{https://languagechange.org/}}, therefore several datasets have been proposed for different languages \cite{schlechtweg2020semeval2020}. 
This work take place in the context of DIACR-Ita \cite{diacrita_evalita2020} at EVALITA 2020 \cite{Evalita2020}, which sets the task for the Italian language in a fully unsupervised fashion. From DIACR-Ita we received 18 target words\footnote{'egemonizzare', 'lucciola', 'campanello', 'trasferibile', 'brama', 'polisportiva', 'palmare', 'processare', 'pilotato', 'cappuccio', 'pacchetto', 'ape', 'unico', 'discriminatorio', 'rampante', 'campionato', 'tac', 'piovra'}, and two time-specific and preprocessed Italian corpora, namely \textit{T0} and \textit{T1}, which include part-of-speech tagging and lemmatization information.


We present two perspectives to approach the problem,  regarding how we represent target words and estimate the lexical-semantic change across datasets. (1) uses the POS distribution of target words as representation, and employee an ensemble of distance measures for the estimation. (2) uses the target words neighbor similarities as representation and one (of two proposed) similarity measure for estimation.


The following three sections describe the previous works, modeling, and results we obtained using these approaches. Following that, section \ref{discussion_section} (Discussion) focuses on examine the second approach to illustrate the impact of the $k$ parameter in similarity measures and the discriminatory performance of our embedding-based model.

\section{Related works}

Previous works have employed similar approaches to address the unsupervised lexical-semantic-change task, mostly for the English language \cite{schlechtweg2020semeval2020,asgari2020unsupervised,schlechtweg2020semeval}. Our first approach follows the idea of ``syntactic models" \cite{kulkarni2014statistically}, which supposes that some semantic changes could imply a new syntactic functionality, such as acquiring a new part-of-speech category, as \newcite{kulkarni2014statistically} exemplify: the word ``apple" increased his use as a proper name in the '80s.

On the other hand, our second approach follows the idea of ``embedding-based models" \cite{kulkarni2014statistically,hamilton-etal-2016-cultural,shoemark-etal-2019-room}, which compares word vector representations from each period using an aligned space, which can be computed either globally (for the full model) or locally (only for a target words).
A common strategy for local aligning is to perform a new transformation representing the target words (the same from different spaces) through neighborhood structures, under the assumption that independent training of embedding algorithms on comparable corpora will still produce similar neighborhood structures \cite{kulkarni2014statistically}. 



Our second approach align the space locally using the nearest neighbors of target words as shared feature.

\section{Methodology}
In this section we provide a detailed description of our systems, each of them composed of two stages, the model and the voting scheme.

\subsection{Models}

We represented the target words as vectors for each time of period using two perspectives that originate our submitted systems: the POS-model and the embedding-model. The word representations are comparable across spaces, and serve to estimate the lexical semantic change through similarity and distance measures, from which we finally predict the changing words using thresholds and voting schemes.

\par \textbf{POS-model:} we simply analyzes the Part-Of-Speech distribution as the relative frequency over the datasets taking the top 4 most common POS-tags, namely ADJ, NOUN, PROPN and VERB.
The produced four-dimensional vector pairs are then used
to assess the lexical semantic change of each target word from the perspective of their Euclidean, Manhattan and Cosine distances\footnote{we noticed that at this point \newcite{kulkarni2014statistically} uses Jenssen-Shannon divergence measure}.

\par \textbf{Embedding-model:}
We lowercase and concatenate each word form with its corresponding POS to build embedding models for each dataset $T$, namely $T0$ and $T1$. 
Specifically, we used Word2Vec models\cite{mikolov2013distributed} with the CBOW version from gensim%
\footnote{https://radimrehurek.com/gensim/models/word2vec.htm} 
with the following parameters: size of 256, window of 5, min\_count of 3. Then we take the common vocabulary of both  $V_c = V_{(TO)} \cap V_{(T1)}$, 
and use it to constraint the set of top $k$ nearest neighbors of the target word only from $T0$%
\footnote{Unlike  \newcite{hamilton-etal-2016-cultural} that takes the top-$k$ neighbors from each model and union them ($N_k=N_k^{T0} \cup N_k^{T1}$).},
i.e., $N_k=\{n_1,n_2...n_k\}, n_k \in V_c$, 
to build the representation of the target word for each space based on its neighbor proximity, i.e. $\vec{W}^{T} = [cos\_sim(\vec{w},\vec{n_k})| n_k \in T]$, 
and estimate the lexical semantic change using the following two formulas%
\footnote{ \newcite{hamilton-etal-2016-cultural} only uses cosine distance.}:
\begin{equation}
\text{\textit{avg.abs.diff}} = Avg(|\vec{W}^{T0} - \vec{W}^{T1}|)
\label{avg.abs.diff}
\end{equation}
\begin{equation}
\text{\textit{cosine\_similarity}} = cos\_sim(\vec{W}^{T0},\vec{W}^{T1})
\label{cos_sim}
\end{equation}
The average of absolute point-wise differences (avg.abs.diff for short) works under the assumption that the neighbors a non-changing word preserves their relative distance each other across diachronic representations. Therefore, the value of this measure increases according to the lexical semantic change a target word underwent. In our submission we used $k=10$.

\subsection{Threshold and voting schemes}
Given that DIACR-Ita is an unsupervised task we experiment with different threshold and voting schemes to aggregate the measure ranks and determine which target words have underwent a lexical semantic change. As a result, we propose three voting schemes from which we derive our results. 

\par \textbf{System1: Upper-third of distance ranks (used for POS model):} we sorted the target words in descending order and rank their positions according to the Euclidean, Manhattan and Cosine distances. We then sum all these ranks and sort in descending order again. Finally we label the first upper-third part of this list as changing words.

\par \textbf{System2: Half intersection (used for the embedding model):} We sort the target words in descending and ascending order for the lineal-difference scores (\ref{avg.abs.diff}) and the cosine-similarity (\ref{cos_sim}) respectively. Then we take the top 50\% of each group, and intersect them to obtained the words that we predicted as changing words.

\par \textbf{System3: Union of Upper-third and Half intersection:} This is just the union of results from System1 and System2.

\begin{figure*}[!ht]
\begin{centering}
\includegraphics[width=1\linewidth]{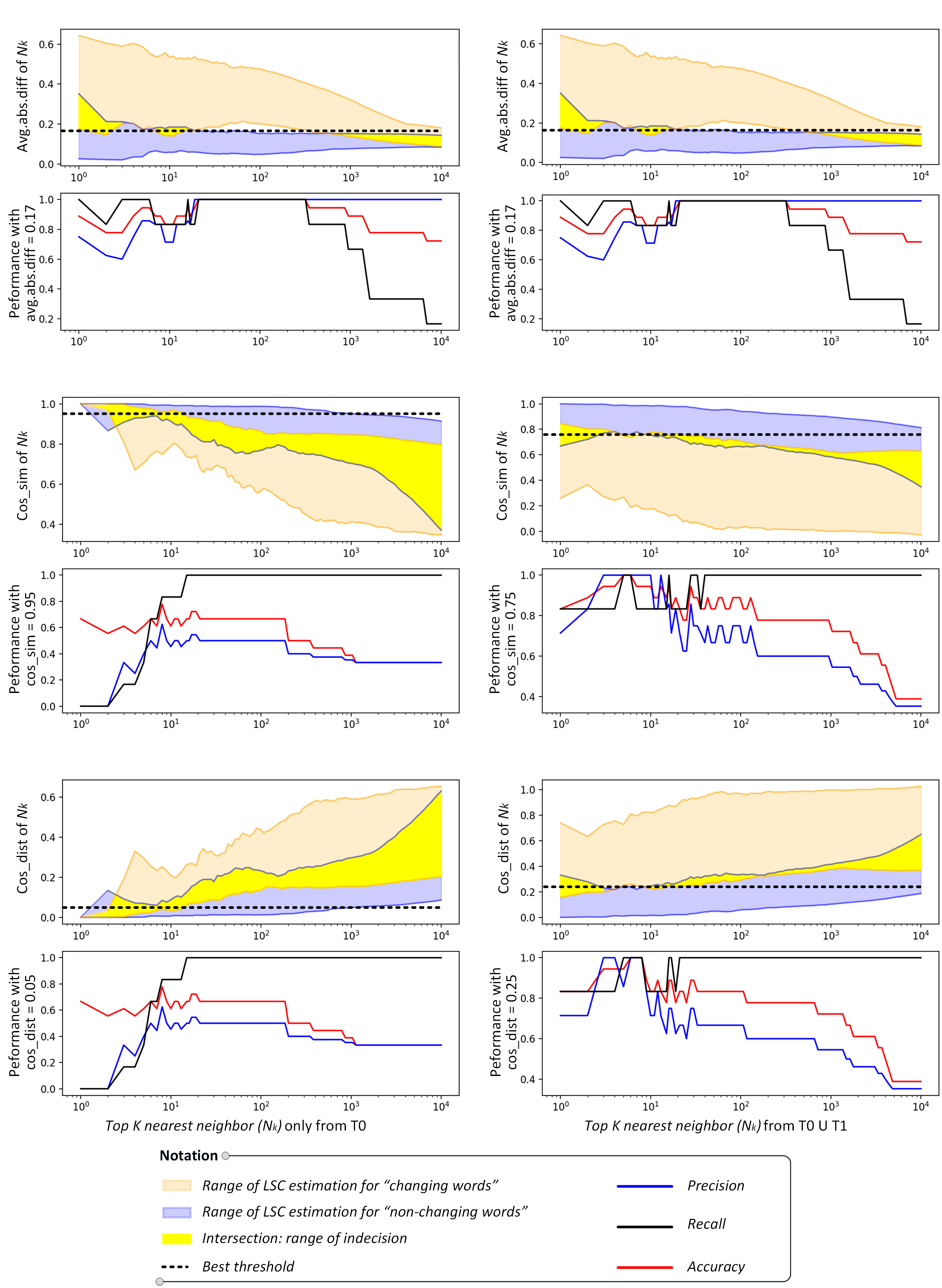}
\par\end{centering}
\protect\caption{ 
Analysis of estimation ranges of lexical semantic change by neighbor-based distributional models using several measures, and two aggregation methods: only from T0 (at left) and the union of T0 and T1 (at right).
} 
\label{fig:analysis}

\end{figure*}

\section{Results}
Table \ref{results_table} summarize the results we obtained during the competition. One can see that the system3 which combine system 2 and 3 also combine its false positive results while removing the False negative ones.  We officially ranked third place with the System1, which in spite of exhibit equal results than System3, is much simpler. We also made error analysis over the system 1 for the case of ``polisportiva" at Table \ref{polisportiva_table}, the results show that there is a large difference in the POS usage of ``polisportiva" across the time periods, NOUN and PROPN seems to invert their distribution usage. We also made the code\footnote{\url{https://github.com/ajason08/evalita2020_diacrita}} publicly available for the systems reproduction.

\begin{table}[h]
\begin{center}
\begin{tabular}{|c | c |p{2.18cm}|p{2.28cm}|}
\hline \bf S (\#)& \bf Acc. & \bf False positive & \bf False negative \\ \hline
\textbf{1} & 0.88 &  polisportiva &  rampante \\ 
\textbf{2} & 0.83 &  egemonizzare &  lucciola, ape \\ 
\textbf{3} & 0.88 &polisportiva, egemonizzare&  -- \\ \hline
\end{tabular}
\end{center}
\caption{ Submission results using Accuracy }
\label{results_table}
\end{table}

\begin{table}[h]
\begin{center}
\begin{tabular}{|l|c|c|c|c|}
\hline \bf  Corpus & \bf  ADJ & \bf NOUN & \bf PROPN & \bf VERB \\ \hline
\textbf{T0} & 0.04&	0.18&	0.76&	0.02 \\
\textbf{T1} & 0.02&	0.61&	0.34&	0.02 \\
\hline
\end{tabular}
\end{center}
\caption{POS usage of ``polisportiva" over the time periods}
\label{polisportiva_table}
\end{table}



\section{Discussion: Post-evaluation analysis}\label{discussion_section}
In this section we employee the gold-standard labels of the target words to analyze at Figure \ref{fig:analysis} the capabilities of our neighbor-based embedding-model using several settings.
To this end, we divide the Figure 1 into vertical and horizontal views.
The vertical view defines 3 groups (from top to bottom), that serves to compare the three proposed measures to estimate the lexical semantic change,
namely the average of absolute differences, cosine similarity and cosine distance.
At the same time, the horizontal view serves to compare the strategy of only use T0 (at left), versus the union of T0 and T1 (at right), to define the top nearest neighbors $N_k$.

Next, each of the charts shows an analysis of the model for the given measure across the $k$ parameter.
The area charts represent by color regions the ranges that discriminate the lexical semantic change of target words: ``changing words" (orange region) and ``non-changing words" (purple region). The yellow region in the middle marks the intersection of these ranges, thus, words falling into the yellow region are difficult to estimate, according to the used measure. We also identified the threshold that best discriminate changing and non-changing target words, and draw a dashed line at that point. 
On the other hand, the line charts throw light on all the possible performance that the model could obtain by changing the $k$ parameter while using the best possible discriminator threshold.

These results suggest that the ``average of absolute difference" is the best proposed measure because it obtains a better performance for a larger number of $k$ values as displayed in the line charts. Moreover, the ``average of absolute difference" offers a larger range for possible discriminator thresholds (as shown in the area charts), and it is tolerant to the $N_k$ election, since it remains almost unchanged while using either the union of T0 and T1, or only T0.
One can also note that the area charts for the cosine similarity versus cosine distance mirror each other, as expected, and their performance is the same when using $N_k$ only from T0 (at left), but slightly differ when using $N_k$ as the union of T0 and T1 (at right).



\section{Conclusion}
We tackle the problem of unsupervised lexical semantic change on two time-specific datasets for 18 target words in Italian language. Our two approaches focus on the representation of target words across the provided diachronic datasets, they use part-of-speech usage and nearest neighbors respectively, and a number of measures between these representation to estimate the lexical semantic change. Then, this estimation serves to decide which target words underwent a change by the use of proposed threshold and voting schemes. Afterward, in the last part of this work, we analyzed the nearest neighbor model through the impact of deciding the $k$ parameter and the similarity measure that estimates the lexical semantic change. Our results for the DIACR-Ita datasets suggest that the estimations of ``the average of absolute differences" measures have a better performance for a larger number of $k$ values than the cosine similarity and the cosine distance used in \newcite{hamilton-etal-2016-cultural}. 

As for future work, we plan to investigate different mechanism for deciding the threshold, and explore other diachronic datasets for other languages such as English, German and Spanish. We also believe that further experiments on a larger number of target words will benefit the reliability of models to judge the lexical semantic change in an unsupervised fashion.




\section*{Acknowledgments}

The authors thank CONACYT for the computer resources provided through the INAOE Supercomputing Laboratory's Deep Learning Platform for Language Technologies.

\bibliographystyle{acl}
\bibliography{references}

\end{document}